\documentclass{fcs}
\usepackage{bm}
\usepackage{url}
\usepackage{multirow}
\usepackage{array}
\usepackage[table]{xcolor}
\usepackage[font=small,labelfont=bf]{caption}
\volumn{ }
\doi{ }
\articletype{RESEARCH~ARTICLE}
\copynote{{\copyright} Higher Education Press and Springer-Verlag Berlin Heidelberg 2012}
\ratime{Received month dd, yyyy; accepted month dd, yyyy}
\email{$fanws@lamda.nju.edu.cn$}
\title{Exploring Dark Knowledge under Various Teacher Capacities and Addressing Capacity Mismatch}
\author{Wen-Shu \MakeUppercase{Fan} \xff $^{1,2}$, Xin-Chun \MakeUppercase{Li} $^{1,2}$, De-Chuan \MakeUppercase{Zhan} $^{1,2}$}
\address{{1\quad School of Artificial Intelligence, Nanjing University, Nanjing 210023, China}\\
{2\quad National Key Laboratory for Novel Software Technology, Nanjing University, Nanjing 210023, China}}

\begin{document}
\maketitle
\setcounter{page}{1}
\setlength{\baselineskip}{14pt}

\begin{abstract}
Knowledge Distillation (KD) could transfer the ``dark knowledge" of a well-performed yet large neural network to a weaker but lightweight one. From the view of output logits and softened probabilities, this paper goes deeper into the dark knowledge provided by teachers with different capacities. Two fundamental observations are: (1) a larger teacher tends to produce probability vectors with lower distinction among non-ground-truth classes; (2) teachers with different capacities are basically consistent in their cognition of relative class affinity. Through abundant experimental studies we verify these observations and provide in-depth empirical explanations to them. We argue that the distinctness among incorrect classes embodies the essence of dark knowledge. A larger and more accurate teacher lacks this distinctness, which hampers its teaching ability compared to a smaller teacher, ultimately leading to the peculiar phenomenon named "capacity mismatch". Building on this insight, this paper explores multiple simple yet effective ways to address capacity mismatch, achieving superior experimental results compared to previous approaches.
\end{abstract}


\section{Introduction} \label{sec:intro}
\noindent Deploying large-scale neural networks on portable devices with limited computation and storage resources is challenging~\cite{ModelCompression}, and efficient architectures such as MobileNets~\cite{MobileNet, MobileNetV2} and ShuffleNets~\cite{ShuffleNet, ShuffleNetV2} have been designed for lightweight deployment. However, the performances of these lightweight networks are usually not comparable to the larger ones. Commonly, second learning~\cite{SecondLearn-Rules, SecondLearn-NeC4.5} or knowledge distillation (KD)~\cite{DoDeep, KD, TNNLS-Collaborative} could be utilized to transfer the knowledge of a more complex and well-performed network (i.e., the teacher) to the smaller ones (i.e., the student). Both are manifestations of pre-learning transfer, with KD focusing more on extracting the dark knowledge from the teacher model. However, the dark knowledge in KD is still a mystery that has attracted lots of studies~\cite{KD, KDPrivileged, TowardsKD, StasticalKD}, and their goal is to answer the following question: what's the knowledge that the teacher provides and why they are effective in KD? 

In the original KD method~\cite{KD}, the student aims to mimic the teacher's behavior by minimizing the Kullback-Leibler (KL) divergence between their output probabilities. That is, the logits and softened probabilities, i.e., the inputs to the final softmax operator and the corresponding outputs, are the specific knowledge transferred in KD. With the development of KD methods, the output-level knowledge has been extended to various types~\cite{KD-Survey-IJCV}, including the intermediate features~\cite{FeatKD-FitNet, FeatKD-AT, FeatKD-NST, FeatKD-AB, FeatKD-PKT, TNNLS-Feature}, the sample relationships~\cite{ReKD-CC, ReKD-CRD, ReKD-Gift, ReKD-Graph, ReKD-Metric, ReKD-SP, ReKD-Refilled}, the parameters~\cite{ParamKD-KR, ParamKD-L2SP}, and the collaborative or online knowledge~\cite{TNNLS-Collaborative, TNNLS-Online} etc. Among these types, the output logits of neural networks are much easier to visualize, analyze, and understand. Therefore, we focus on the original KD~\cite{KD} and aim to understand the dark knowledge (i.e., the logits and softened probabilities) provided by the teachers. Unlike previous studies, we majorly study the output-level dark knowledge provided by teachers {\it with various capacities}, which receives little attention in previous studies. We first present two significant observations: (1) an over-confident teacher tends to produce probability vectors that are less distinct between non-ground-truth classes; (2) teachers with different capacities are basically consistent in their cognition of relative class affinity. The first observation tells the difference between dark knowledge provided by teachers with different capacities, while the second observation shows the consistency between them.

This paper first explains the reasons for the first observation. Larger teachers generally have powerful feature extractors, making the features of the same class more compact and the features between classes more dispersed. Hence, more complex teachers are over-confident and assign a larger score for the ground-truth class or less varied scores for the non-ground-truth classes. If we use a uniform temperature to scale their logits, the probabilities of non-ground-truth classes are less distinct~~\cite{ATS}, further making the distillation process ineffective. This explains the peculiar phenomenon named ``capacity mismatch"~\cite{KDEfficacy, TAKD, SCKD, ResKD-NC, ResKD-TIP, ATS} in KD that a larger and more accurate teacher does not necessarily make better students.
Fortunately, the second observation ensures that the dark knowledge from teachers with various capacities is basically consistent in the class relative probabilities. We first provide several definite metrics to verify the observation, including the rank of set overlap, Kendall's $\tau$, and the Spearman correlation. These metrics are irrelevant to the absolute probability values between classes, which are appropriate for measuring the correlation of relative class affinities. We also present the two observations together via the group classes in CIFAR-100~\cite{cifar} and a constructed group classification task based on Digits-Five~\cite{DaNN}.

These two observations imply that complex teachers know approximately the same as smaller teachers on relative class affinities, while their absolute probabilities are not discriminative between non-ground-truth classes. Hence, to improve the quality of the dark knowledge provided by complex teachers, an intuitive way is to enlarge the distinctness between non-ground-truth classes. We propose several simple yet effective methods to enlarge the distinctness of non-ground-truth class probabilities, which could make the distillation process more discriminative. Abundant experimental studies verify that the proposed methods could address the capacity mismatch problem effectively.

We summarize our contributions as two aspects: (1) showing novel insights about the dark knowledge provided by teachers with various capacities, including their distinctness on absolute class probabilities and consistency in relative class affinities; (2) addressing the capacity mismatch problem in KD by proposing multiple simple yet effective methods, which are verified by abundant experimental studies.

\section{Related Works} \label{sec:relate}
Our work is closely related to the dark knowledge and capacity mismatch problem in KD.

\subsection{Dark Knowledge in KD}
Quite a few works focus on understanding and revealing the essence and effects of ``dark knowledge" in KD, including the empirical studies and theoretical analysis~\cite{ImproveKD, KDWide, StasticalKD, TowardsKD, KDLSR, KDSemiParametric, KDBiasVar, KDEfficacy, KDPrivileged, GeneralizeBound}. \cite{KDPrivileged} unifies the dark knowledge with privileged information, and \cite{KDConcept} attributes the dark knowledge in vision tasks to the task-relevant and task-irrelevant visual concepts. The dark knowledge in KD is also related to label smoothing (LS)~\cite{KDLSR, WhenLSHelp, ISLS, Boosting-intra-class}. Some works decompose the dark knowledge of teachers and aim to understand their effectiveness correspondingly. \cite{BAN} decomposes the dark knowledge into two parts, explaining the teacher's ground-truth/non-ground-truth outputs as importance weighting and class similarities. \cite{ImproveKD} decomposes the dark knowledge into three parts including universal knowledge, domain knowledge, and gradient rescaling. The bias-variance decomposition is also utilized to analyze the property of KD~\cite{StasticalKD, KDBiasVar}. \cite{TNNLS-Customized} utilizes the standard deviation of the secondary soft probabilities to reflect the quality of the teachers' dark knowledge. Recently, \cite{Diffusion} proposed an innovative perspective, suggesting that the knowledge of the teacher stems from containing less noise. The most related work~\cite{ATS} decomposes the KD into correct guidance, smooth regularization, and class discriminability, which provides novel insights about the key information in the dark knowledge. Few works have examined the relationship of dark knowledge with teacher capacities, and this paper provides detailed observations and analysis about this.

\subsection{Capacity Mismatch in KD}
An intuitive sense after the proposal of KD~\cite{KD} is that larger teachers could teach students better because their accuracy is higher. However, there exists a peculiar phenomenon named capacity mismatch that excellent teachers can't completely teach the smaller students well. ESKD~\cite{KDEfficacy} first points out the phenomenon and points out that teacher accuracy may poorly predict the student’s performance, i.e., more accurate neural networks don't necessarily teach better. However, no research so far has provided an accurate explanation or even a quantitative definition for this phenomenon. Previous works have only been made to propose solutions from various perspectives. TAKD~\cite{TAKD} solves this problem by introducing an intermediate-sized network (i.e., the teacher assistant) to bridge the knowledge transfer between networks with a large capacity gap. SCKD~\cite{SCKD} formulates KD as a multi-task learning problem with several knowledge transfer losses. ResKD~\cite{ResKD-NC, ResKD-TIP} utilize the ``residual" of knowledge to teach the residual student, and then take the ensemble of the student and residual student for inference. DIST~\cite{StrongTeacher} finds that the discrepancy between the student and a stronger teacher may be fairly severe, which disturbs the training process, and they propose a ranking-based loss as the solution. \cite{CheckpointKD} advocates that an intermediate checkpoint will be more appropriate for distillation. \cite{FS-Strong} explores the strong teachers in few-shot learning. \cite{TPAMI-Gap} studies the effect of capacity gap on the generated samples in data-free KD. Recently, \cite{LSKD} demonstrated that sharing the same temperature between the teacher and student leads to a logit shift, which is a side-effect in KD. One of our methods is closely related to~\cite{ATS} that proposes Asymmetric Temperature Scaling (ATS) to make the larger networks teach well again. An improved version named Instance-Specific ATS (ISATS) is proposed. Aside from this method, we also propose other simple yet effective solutions from different aspects.

\section{Background and Preliminaries} \label{sec:bg}
\noindent We follow the basic notations as introduced in~\cite{ATS}. Specifically, we consider a $C$-class classification problem with $\mathcal{Y}=[C]=\{1,2,\cdots,C\}$. We denote the ``logits" as the output scores of a sample $\mathbf{x}$ before applying the softmax function, which is represented as $\mathbf{f}(\mathbf{x}) \in \mathbb{R}^C$. Correspondingly, the softened probability vector is denoted as $\mathbf{p}(\mathbf{x};\tau)=\text{SF}(\mathbf{f}(\mathbf{x});\tau)$:
\begin{equation}
    \text{SF}(\mathbf{f}(\mathbf{x});\tau)=\exp \left(\mathbf{f}(\mathbf{x}) / \tau\right) / Z(\tau),
\label{softmax prediction}
\end{equation}
where $Z(\tau) = \sum_{i=1}^{C}\exp(f_{i}(\mathbf{x})/\tau)$. $f_{i}(\mathbf{x})$ is a scalar which refers to the logit of the $i$-th class in $\mathbf{f}(\mathbf{x})$. $\mathbf{f}_y$ and $\mathbf{p}_y$ denote the ground-truth class's logit and probability, while $\mathbf{g}=[\mathbf{f}_c]_{c\neq y}$ and $\mathbf{q}=[\mathbf{p}_c]_{c\neq y}$ represent the vector of non-ground-truth classes' logits and probabilities. $y$ denotes the ground-truth class.

The most standard KD~\cite{KD} contains two stages of training. The first stage trains complex teachers, and then the second stage transfers the knowledge from teachers to a smaller student by minimizing the KL divergence between softened probabilities. Usually, the loss function during the second stage (i.e., the student's learning objective) is a combination of cross-entropy loss and distillation loss:
\begin{equation}
    \ell = -(1-\lambda)\log \mathbf{p}^s_y(1) -\lambda \tau^2 \sum_{c=1}^C \mathbf{p}^t_{c}(\tau)\log \mathbf{p}^s_c(\tau), \label{eq:loss}
\end{equation}
where the upper script ``t''/``s'' denotes ``teacher''/``student'' respectively. We utilize $
\lambda\in[0,1]$ as a hyperparameter to balance the weights between the two training stages. The dark knowledge in the above distillation equation is the teacher's label, i.e., $\mathbf{p}^t(\tau)$. In this paper, we aim to study the influence of teacher capacity on the teacher's label. That is, we majorly compare the relationships between $\mathbf{p}^{t_{\text{large}}}$ and $\mathbf{p}^{t_{\text{small}}}$ which are provided by a larger and a smaller teacher, denoted as $t_{\text{large}}$ and $t_{\text{small}}$, respectively.

The process of KD has the effect of improving the student network's performance when compared with its independent training. An explanation for this improvement is that the dark knowledge contained in larger teachers could help the student better capture the semantic information. Commonly, the training and test accuracy of $t_{\text{large}}$ will be higher than that of $t_{\text{small}}$ because the larger teacher has a huge capacity to capture more information. Hence, to further improve the performance of the student network, a natural idea is replacing the smaller teacher (i.e., $t_{\text{small}}$) with a larger one (i.e., $t_{\text{large}}$). Frustratingly, this leads to a degradation in the performance of the student network. This phenomenon is named the ``capacity mismatch'' in KD, which is still counter-intuitive, surprising, and unexplored as declared in previous studies~\cite{SCKD, TAKD, ResKD-TIP}. This paper will propose multiple simple yet effective methods to tackle the performance degradation problem.

\begin{figure*}[tbp]
    \centering
    \includegraphics[width=\linewidth]{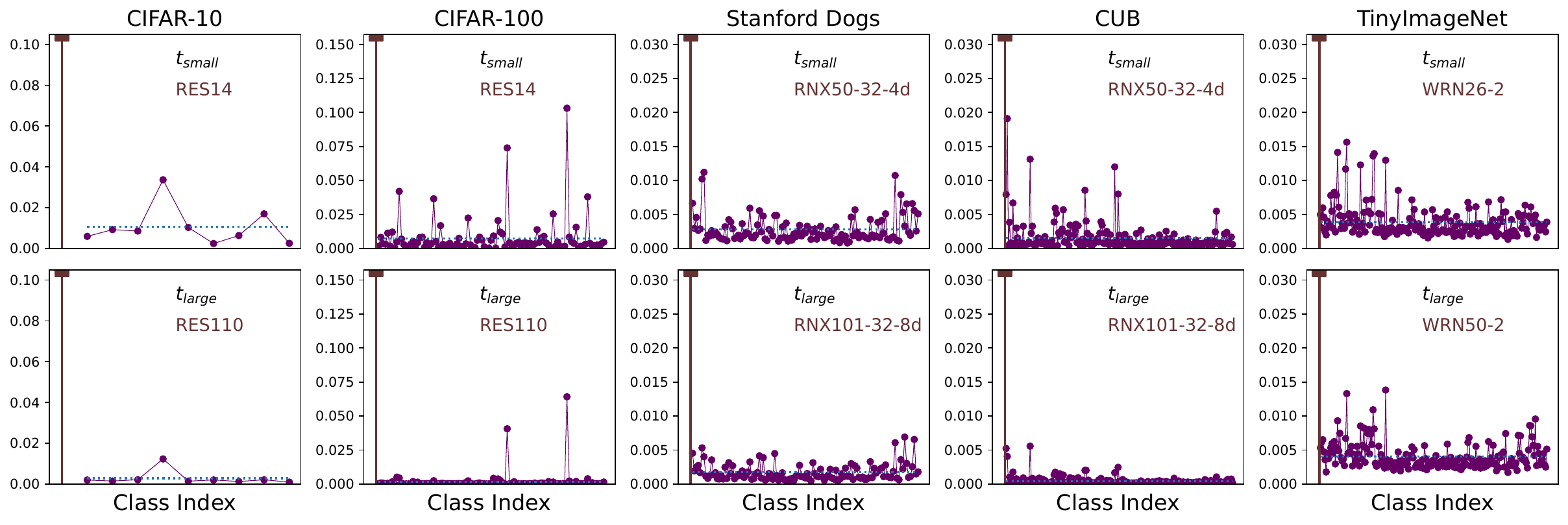}
    \caption{Visualization of softened probability vectors. (Up) Results provided by smaller teacher on various datasets; (Bottom) Results provided by larger teacher on various datasets. Each column’s two rows represent the same sample with the same temperature. The first bar shows the ground-truth class while others are non-ground-truth classes. Two observations: (1) the non-ground-truth class probabilities shown in the second row are less varied; (2) the relative class probabilities are nearly the same between the first and second rows.}
    \label{fig:obser}
\end{figure*}

\section{Observations and Explanations}
\noindent We first present two observations about the influence of teacher capacities on dark knowledge. Then, we provide explanations and metrics to better understand the findings.
	
\subsection{Observations} \label{sec:obser}
An intuitive method to reflect the relationship of dark knowledge produced by teachers with various capacities is visualization. The first step is training teacher networks with different capacities on several benchmarks. Specifically, we train ResNet14 and ResNet110~\cite{ResNet} on the CIFAR-10/CIFAR-100 dataset~\cite{cifar}, ResNeXt50-32-4d and ResNeXt101-32-8d~\cite{ResNeXt} on the Stanford Dogs~\cite{dogs} and CUB~\cite{CUB} dataset, and WideResNet26-2 and WideResNet50-2~\cite{WideResNet} on the TinyImageNet~\cite{TinyImageNet} dataset. We maximize the size difference between the two models on each dataset to make the experimental effects more pronounced. Then, for each dataset, a random sample is sampled from the first class in the training set and the softened probability vectors produced by different teachers are visualized. Fig.~\ref{fig:obser} shows the results, where ``RES", ``RNX", and ``WRN" are abbreviations for corresponding networks. The utilized temperature is common across the subfigures. Comparing the first and second rows, it is obvious that larger teachers provide less varied probabilities for non-ground-truth classes. Although the variance of non-ground-truth class probabilities differs a lot between teachers, the relative probability values seem to be consistent. For example, the several top-$K$ bars between the two rows lie nearly in the same classes. We conclude the two fundamental observations as follows: {\it (1) larger teachers tend to produce probability vectors that are less distinct among non-ground-truth classes; (2) teachers with different capacities are basically consistent in their cognition of relative class affinity.} To further support these two observations, we provide explanations and statistical measures in the following.

\begin{figure}[tbp]
    \centering
    \includegraphics[width=\linewidth]{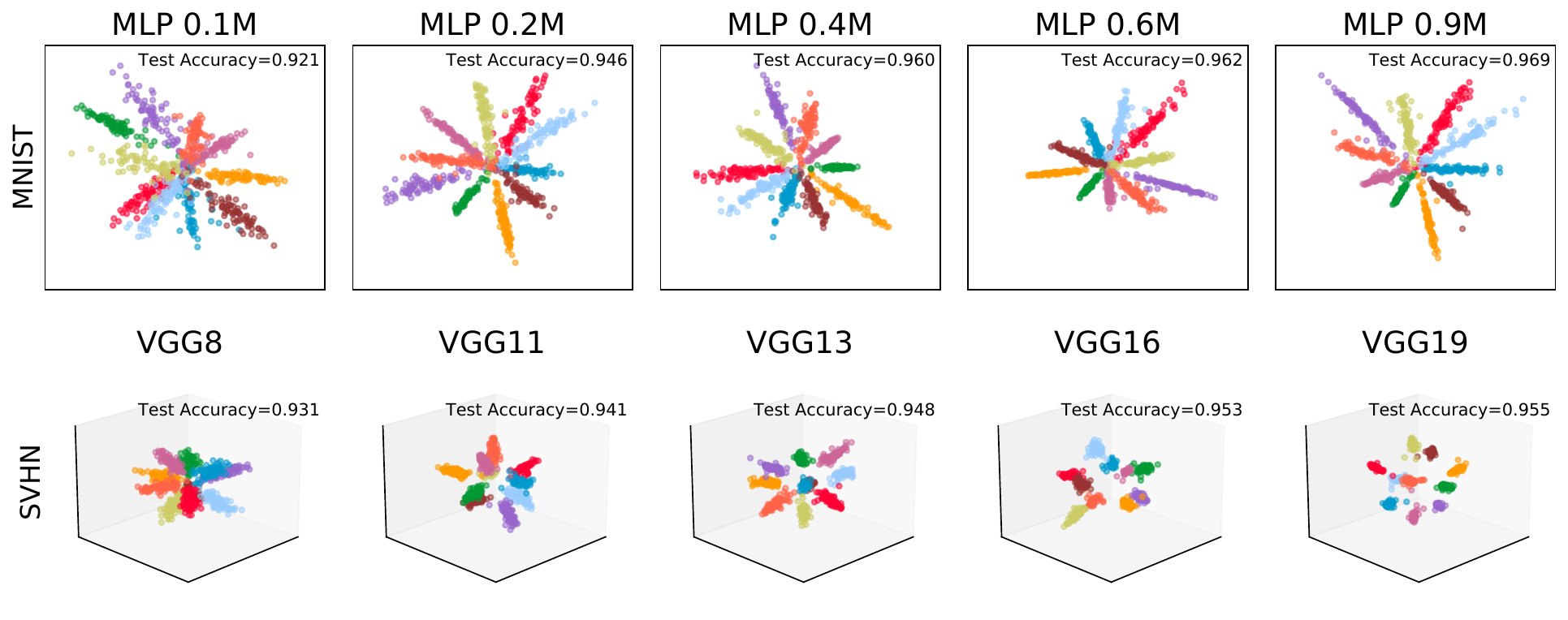}
    \caption{Visualization of extracted features by teachers with increasing capacities. The first row shows the extracted 2-dim features on MNIST, and the second shows the 3-dim features on SVHN. The accuracy on the test set is also reported.}
    \label{fig:vis}
\end{figure}

\subsection{The Variance of Non-ground-truth Class Probabilities} \label{sec:small-dv}
The explanations for the first observation are provided step by step. 

\textbf{Low-dimensional features.} First, we train MLP networks on MNIST~\cite{mnist} and train VGG~\cite{VGG} networks on SVHN~\cite{Svhn}. These two datasets are relatively simpler classification tasks, and we set the final feature dimensions as 2 and 3 correspondingly for better visualization. We do not use the bias parameters in the last layer. We progressively enlarge the depth and width of MLP networks, and use the total number of parameters to denote the network capacity. The capacity of MLP networks ranges from 0.1M to 0.9M. The VGG capacities are determined by the number of layers, which ranges from 8 to 19. We do not use ReLU~\cite{relu} activation for the final extracted features. These two datasets have 10 types of digits to identify, and the extracted features are scattered in Fig.~\ref{fig:vis}. With the capacity increases, the features of the same class are more compact, while the class centers between different classes are more dispersed. Notably, the larger networks do not overfit the training dataset, because the test accuracy shown in the figure still becomes higher. This implies that larger DNNs obtain a higher performance but may extract more compact intra-class features and more dispersed inter-class features. 

\begin{figure}[tbp]
    \centering
    \includegraphics[scale=0.6]{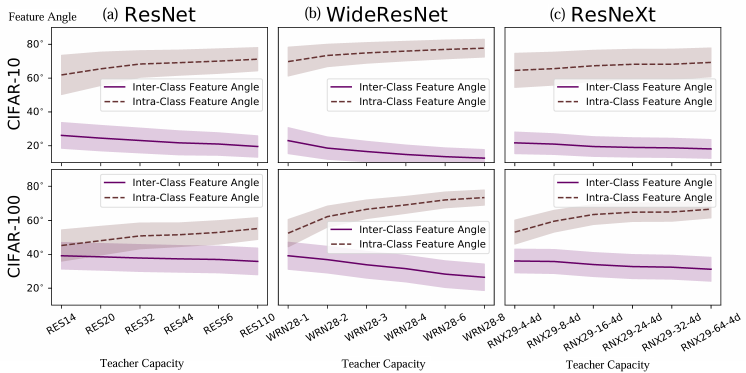}
    \caption{The change of inter-class and intra-class feature angle under different teachers. (a) ResNet teachers; (b) WideResNet teachers; (c) ResNeXt teachers}
    \label{fig:angle}
\end{figure}

\textbf{High-dimensional features.} The above demos show low-dimensional features that may not reflect the change in high-dimensional space. Hence, we extend the findings to CIFAR-10 and CIFAR-100, on which we train a series of ResNet, WideResNet, and ResNeXt networks. Because the high-dimensional features are hard to visualize and the dimensional-reduced plots by T-SNE~\cite{TSNE} are sensitive to hyper-parameters, we instead calculate several metrics to indirectly verify the change of feature compactness. Specifically, we calculate the feature angle as follows:
\begin{equation}
    \mathcal{A}(i, j) = \arccos\left(\frac{\mathbf{h}_i^T\mathbf{h}_j}{\lVert \mathbf{h}_i \rVert \lVert \mathbf{h}_j \rVert}\right), \label{eq:angle}
\end{equation}
where $\mathbf{h}_i$ and $\mathbf{h}_j$ are feature vectors of the $i$-th and $j$-th training sample. Then, we define the inter-class and intra-class set as $\{(i, j)\}_{y_i=y_j, \forall i, j}$ and $\{(i, j)\}_{y_i \neq y_j, \forall i, j}$, respectively. The average and standard deviation of angles in the two sets are plotted in Fig.~\ref{fig:angle}. If the teacher becomes more complex, the inter-class feature angle becomes smaller, while the intra-class angle becomes larger. Notably, we use ReLU~\cite{relu} activation before the classification layer, implying that all of the elements in the extracted features are non-negative. This could already provide some hints for the capacity mismatch phenomenon. As an extreme case, if the features among the same class collapse into a single point, and features among different classes are orthogonal to each other, the inter-class feature angle will be zero and the intra-class one will be 90$^{\circ}$. Although the training and test accuracy will be $100\%$, the softened probabilities may be one-hot labels, which bring no additional information to the student. An illustration and more detailed discussion can be found in Sect.~\ref{sec:dis}. The inter-class and intra-class distance metrics utilized in linear discriminant analysis~\cite{LDA} or unsupervised discriminant projection~\cite{UDC} show the same trend. We omit the display of the results based on these distance metrics because the trends of feature angles are vivid enough to show how the feature compactness changes.

\begin{figure}[tbp]
    \centering
    \includegraphics[width=\linewidth]{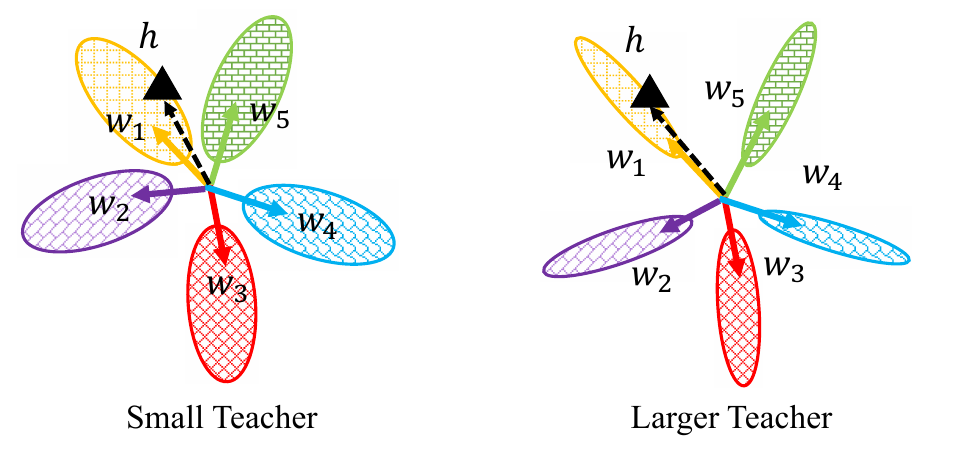}
    \caption{Illustration of the classification layer with different feature compactness. The larger teacher may give a larger ground-truth logit or less varied non-ground-truth logits.}
    \label{fig:teaser-angle}
\end{figure}

\textbf{Visualizaion of classes.} According to the softmax properties in classification problems, the classification weight of each class is pulled towards the features from the same class while pushed away from the features of other classes~\cite{FedRS}. That is, the change tendency of features could partially reflect the tendency of classification weights in the final layer. An ideal case is that the learned classification weights converge to the feature centers. If the intra-class feature angle becomes larger, the feature of a specific sample will be far away from the classification weights of the other classes. Similarly, the feature of a specific sample will be closer to the corresponding class's classification weight. This will lead to a larger logit for the ground-truth class. If the intra-class features are nearly orthogonal, the logits for non-ground-truth classes will be less varied. The illustration can be found in Fig.~\ref{fig:teaser-angle}, where the larger teacher extracts more compact inter-class features and more dispersed intra-class features. The triangle represents a sample feature in the $1$-st class, and the larger teacher tends to give a larger target logit (i.e., $\mathbf{f}_1 = \mathbf{h}^T\mathbf{w}_1$) and less varied non-ground-truth class logits (i.e., $\mathbf{g} = [\mathbf{h}^T\mathbf{w}_c]_{c\in \{2,3,4,5\}}$). A larger target logit or less varied non-ground-truth class logits will both lead to less varied non-ground-truth class probabilities after applying softmax function~\cite{ATS}. 

The step-by-step analysis and visualization provide explanations for the first observation that larger teachers tend to produce probability vectors that are less varied between non-ground-truth classes.

\subsection{The Relative Class Affinities} \label{sec:relative}
\begin{figure}[tbp]
    \centering
    \includegraphics[width=\linewidth]{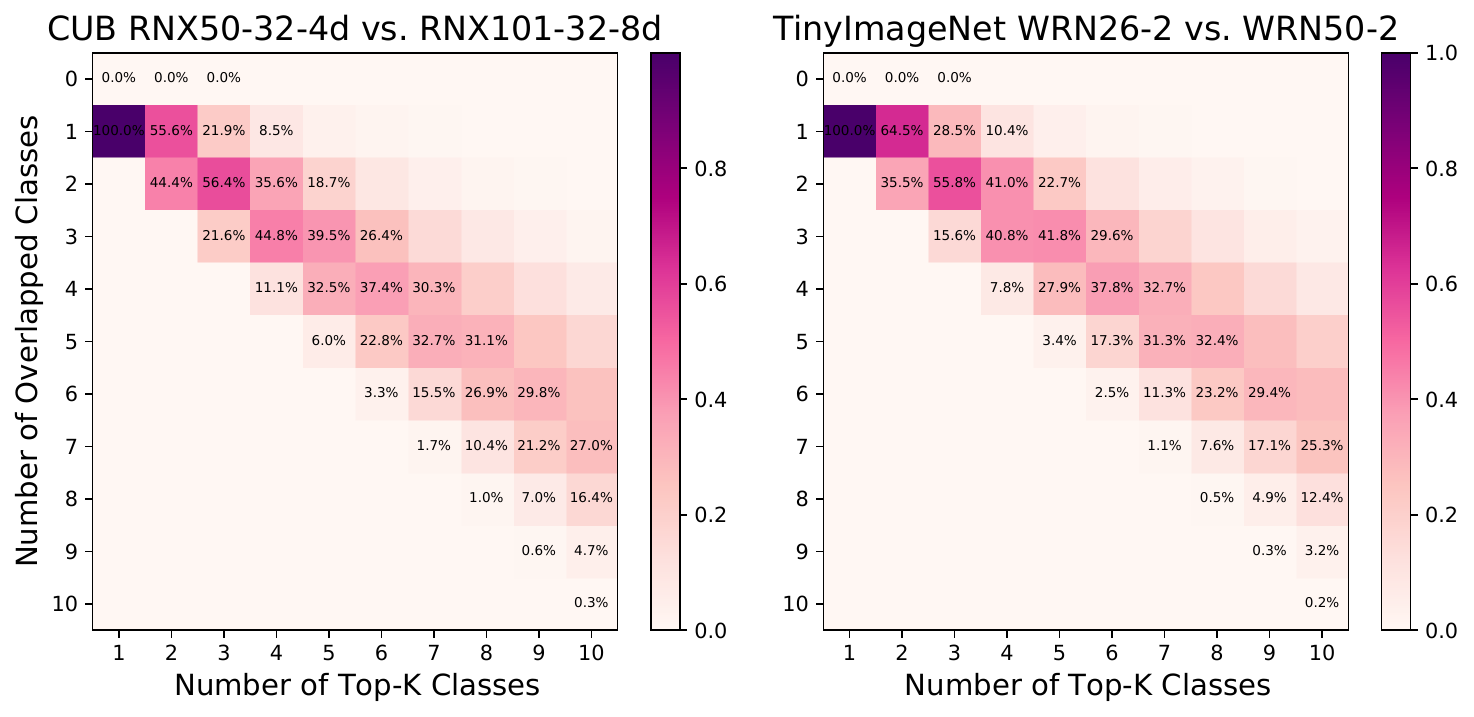}
    \caption{The predicted top-$K$ class set overlap of a smaller and a larger teacher network on two datasets. (Left) CUB dataset; (Right) TinyImageNet dataset}
    \label{fig:set-overlap}
\end{figure}

The second observation in Sect.~\ref{sec:obser} implies that the relative probability values given by different teachers seem to be consistent. Hence, we explore this observation further through several quantitative metrics. The intuitive idea is to explore whether different teachers give the same top-$K$ predictions or not. Specifically, we train a smaller teacher $t_{\text{small}}$ and a larger teacher $t_{\text{large}}$ on the same dataset. Then, we calculate the predicted top-$K$ classes for the $i$-th training sample, and denote the predicted class sets as $\mathcal{C}_{i,K}^{t_{\text{large}}}$ and $\mathcal{C}_{i,K}^{t_{\text{small}}}$, respectively. We denote $|\mathcal{C}_{i,K}^{t_{\text{large}}} \cap \mathcal{C}_{i,K}^{t_{\text{small}}}|$ as the number of overlapped classes. Then we show the distribution of the number of overlapped classes when considering different $K$. Fig.~\ref{fig:set-overlap} displays the results on CUB and TinyImageNet. The percentages in the figure show how many data samples have the number of overlapping classes corresponding to the values of the y-axis. Considering the top-$3$ predicted classes of ResNeXt50-32-4d and ResNeXt101-32-8d on CUB, there are $21.6\%$ training samples that have 3 common predicted classes, and $56.4\%$ training samples that have 2 common predicted classes. If we consider top-$8$ predicted classes, there are nearly $69.4\%$ training samples that have at least 5 common predicted classes. This implies that different teachers indeed have a great deal of consistency in top-$K$ class recognition.

\begin{figure}[tbp]
    \centering
    \includegraphics[width=\linewidth]{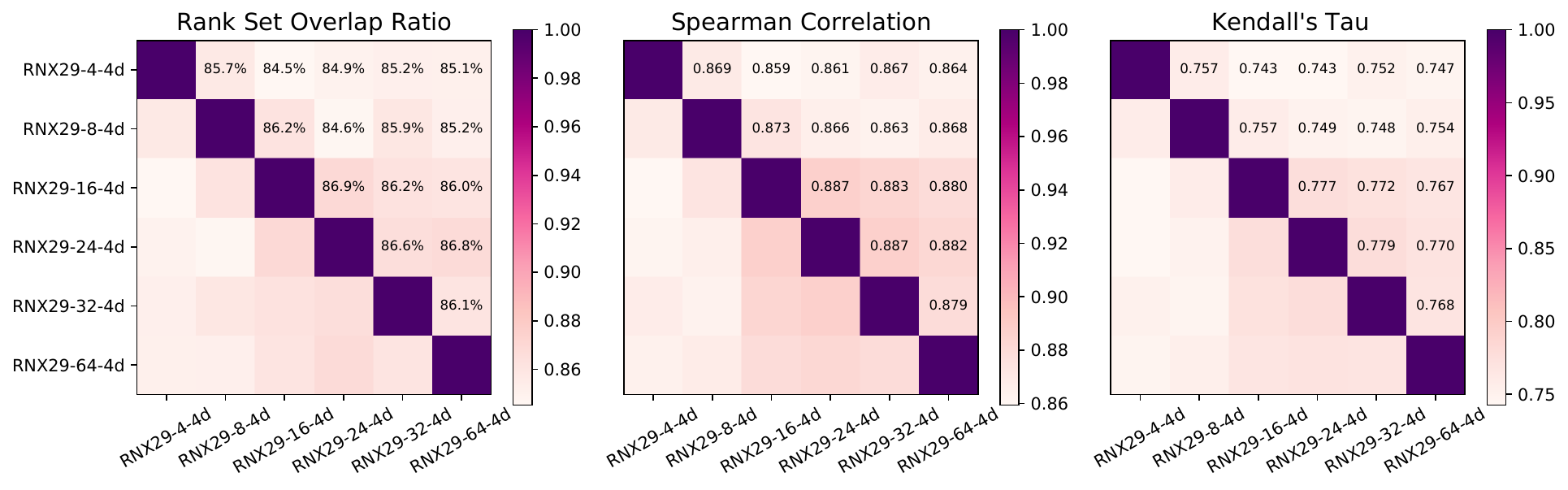}
    \caption{Several magnitude-agnostic metrics that measure the consistency of relative class affinities between teacher networks with various capacities. (Left) Rank Set Overlap Ratio; (Middle) Spearman Correlation; (Right) Kendall's Tau}
    \label{fig:dis}
\end{figure}

Then, we define several specific metrics to further explore the consistency. The first metric is named the rank set overlap ratio, which is calculated as follows:
\begin{equation}
    \textit{s}_{i}(t_1, t_2) = |\mathcal{C}_{i,K}^{t_1} \cap \mathcal{C}_{i,K}^{t_2}| / K, \label{eq:set-overlap}
\end{equation}
where $t_1$, $t_2$ denote teachers. The second metric is the Spearman correlation. It first obtains the rank of all classes in the output logits of two teachers, i.e., $\mathbf{R}_{i}^{t}=\text{argsort}(\mathbf{f}_{i}^{t})$, $t \in \{t_1, t_2\}$. Then, it calculates the Pearson correlation as follows:
\begin{equation}
    \rho_{i}(t_1, t_2) = \frac{\sum_{j=1}^C\left(\mathbf{R}_{i,j}^{t_1} - \overline{\mathbf{R}_{i}^{t_1}}\right)\left(\mathbf{R}_{i,j}^{t_2} - \overline{\mathbf{R}_{i}^{t_2}}\right)}{\sqrt{\sum_{j=1}^C\left(\mathbf{R}_{i,j}^{t_1} - \overline{\mathbf{R}_{i}^{t_1}}\right)^2\sum_{j=1}^C\left(\mathbf{R}_{i,j}^{t_2} - \overline{\mathbf{R}_{i}^{t_2}}\right)^2}}, \label{eq:spearman}
\end{equation}
where $j$ is the index of class and $\overline{\mathbf{R}_{i}^{t}}$ means the average of the elements in $\mathbf{R}_{i}^{t}$. Finally, we calculate Kendall's $\tau$ between $\mathbf{f}_{i}^{t_{1}}$ and $\mathbf{f}_{i}^{t_{2}}$, which directly shows the rank correlation of two teachers. The formulation is:
\begin{equation}
    \tau_{i}(t_1, t_2) = \frac{2}{C(C-1)} \sum_{j_1 < j_2} \text{sg}\left(\mathbf{f}_{i,j_1}^{t_{1}}-\mathbf{f}_{i,j_2}^{t_{1}}\right)\text{sg}\left(\mathbf{f}_{i,j_1}^{t_{2}}-\mathbf{f}_{i,j_2}^{t_{2}}\right), \label{eq:kendall}
\end{equation}
where $j_1$ and $j_2$ show the index of classes. $C$ is the number of classes, and $\text{sg}(\cdot)$ returns 1 or -1 depending on whether the input is positive or not. These three metrics only depend on classes' relative magnitudes and are irrelevant to the absolute values. The metrics are calculated on CIFAR-10 under a series of ResNeXt networks. We set $K=5$ for the rank set overlap metric. The above equations only show the metrics on a single training sample, and we report the average result on 50K training samples. The results are in Fig.~\ref{fig:dis}. Excitingly, these metrics among teachers with different capacities do not vary a lot, and the Spearman correlations are almost all larger than $0.85$. According to the interpretation of Kendall's $\tau$~\cite{KendallTau-Int, LogME}, if the smaller teacher predicts that class $j_1$ is more related to the target class than that of class $j_2$, then the larger teacher has a probability of $0.875$ to give the same relative affinity. Certainly, the output logits of different networks are not completely identical in order, even for the same networks under different training randomness~\cite{Convergent, Rome}. These metrics demonstrate that teachers know approximately the same about relative class affinities.

\subsection{Verification by Group Classification} \label{sec:group}
During training, the cross-entropy loss minimizes the divergence between the predicted posterior distribution and the empirical one-hot label distribution. Although one-hot labels do not capture class relationships, the posterior distribution does. The teacher network’s probability vector, $\mathbf{p}^t$, inherently represents this distribution, serving as an alternative that may help the student learn class affinities. However, since the true posterior distribution for each sample is unknown, it is difficult to ascertain which teacher's label is superior.

\begin{figure}[tbp]
    \centering
    \includegraphics[width=\linewidth]{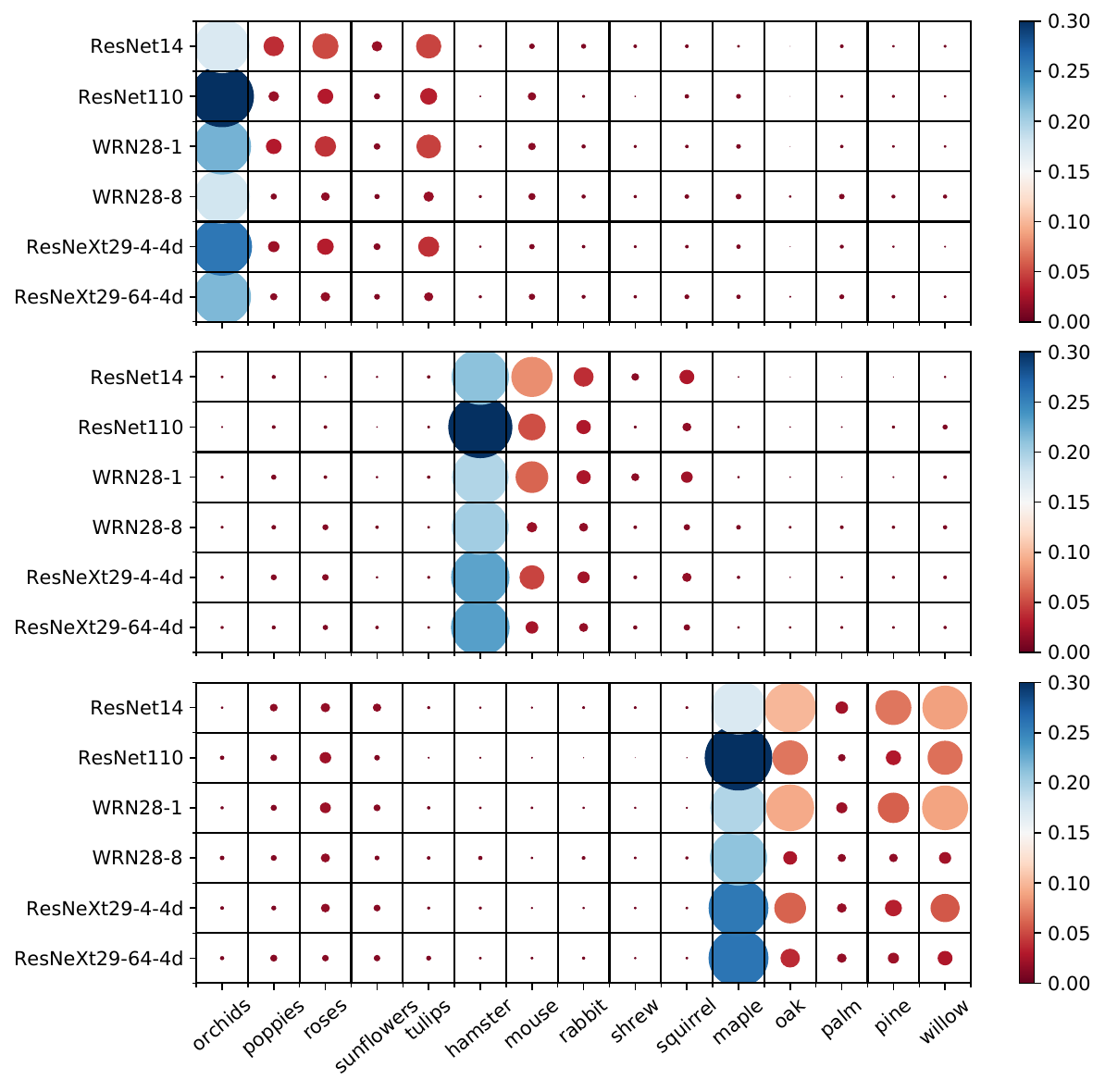}
    \caption{The group classification on CIFAR-100. Each row shows the averaged posterior distribution given by a specific network. Circle size and color map determine the value. The three subfigures show results of fine-grained class ``{\it orchids}", ``{\it hamster}", and ``{\it maple}", respectively, which belong to the superclass ``{\it flowers}", ``{\it small mammals}", and ``{\it trees}". The x-axis shows the $3\times 5=15$ fine-grained classes in the three superclasses.}
    \label{fig:cifar100-group}
\end{figure}

While precise posterior distribution assignment per training sample remains challenging, group classification scenarios offer valuable inter-category priors. CIFAR-100 contains 100 categories, which contain 20 superclasses and 5 classes for each superclass~\footnote{\url{https://www.cs.toronto.edu/~kriz/cifar.html}}. For example, the superclass ``{\it flowers}" contains five fine-grained classes including ``{\it orchids}", ``{\it poppies}", ``{\it roses}", ``{\it sunflowers}" and ``{\it tulips}". One hypothesis is that categories in the same superclass are more similar to each other. We train different teachers (ranges from ResNet14 to ResNeXt29-64-4d) on CIFAR-100 and find that the hypothesis holds true for most superclasses. Fig.~\ref{fig:cifar100-group} presents averaged posterior distribution on three superclasses: ``{\it flowers}", ``{\it small mammals}", and ``{\it trees}". For each superclass, we select its first fine-grained class as the target label, i.e., ``{\it orchids}", ``{\it hamster}", and ``{\it maple}". For each target label, we obtain its corresponding training samples, i.e., 500 samples for each. We calculate the averaged probabilities via $\mathbb{E}_{\mathbf{x},y=c}[\mathbf{p}^t(\mathbf{x};\tau)]$ ($\tau$=4.0) for $c$ being one of the three target classes. The circle size and color map represent the probability value. Key observations: (1) Target classes exhibit stronger affinity to intra-superclass members across all networks; (2) Relative intra-group probabilities maintain cross-architectural consistency. For example, different teacher networks agree that fine-grained classes ``{\it oak}" and ``{\it willow}" are more relevant to the target category ``{\it maple}", while ``{\it palm}" is less relevant. (3) Probability variance decreases with increasing model capacity, evident from diminishing circle size differences in deeper networks. For example, the difference in circle sizes in the last row is much smaller than that in the first row, which behaves similarly in each subfigure.

Then we provide some rules that define the similarity between classes. For example, the rule ``$c_1$: $c_2$, $c_3$, $c_4$, $c_5$" represents that a sample in class $c_1$ is predicted with the following rank: $\mathbf{p}_{c_1} > \mathbf{p}_{c_2} > \mathbf{p}_{c_3} > \mathbf{p}_{c_4} > \mathbf{p}_{c_5}$. If a network gives the rank as ``$c_1$, $c_3$, $c_2$, $c_4$, $c_5$", then the Kendall's $\tau$ between the prediction and the rule is 0.8. To measure the absolute difference in the predictions of non-ground-truth classes, we also calculate the standard deviation of the probabilities, i.e., $\sigma\left(\left[\mathbf{p}_{c_2}, \mathbf{p}_{c_3}, \mathbf{p}_{c_4}, \mathbf{p}_{c_5} \right]\right)$. The following lists the rules defined in CIFAR-100:
\begin{itemize}
    \item \textbf{C.1}: {\it hamster: mouse, rabbit, squirrel, shrew}
    \item \textbf{C.2}: {\it orchids: tulips, roses, poppies, sunflowers}
    \item \textbf{C.3}: {\it maple: oak, willow, pine, palm}
    \item \textbf{C.4}: {\it dolphin: whale, seal, otter, mammals beaver}
    \item \textbf{C.5}: {\it apples: sweet peppers, oranges, pears, mushrooms}
\end{itemize}

These rules are derived from the averaged posterior probability distributions in Fig.~\ref{fig:cifar100-group}. For simplicity, we only consider intra-superclass relationships. We further construct a group classification task using five digit recognition datasets: MNIST, MNIST-M, USPS, SVHN, and SYN. We denote this as Digits-Five, which is often used in domain adaptation tasks~\cite{DaNN} or task relationship estimation~\cite{TaskRelation}. The construction process is simple where we only centralize the data together and obtain a task with $5\times 10=50$ classes. Based on transfer performance analyses and task relation coefficients from existing research~\cite{DaNN, TaskRelation}, we abstract several rules as follows:
\begin{itemize}
    \item \textbf{D.1}: {\it MNIST@c: USPS@c, SVHN@c}
    \item \textbf{D.2}: {\it MNIST@c: MNISTM@c, SYN@c}
    \item \textbf{D.3}: {\it USPS@c: MNIST@c, MNISTM@c, SVHN@c}
\end{itemize}

{\it MNIST@c} represents the $c$-th class in MNIST, and others are defined similarly. Each of the above rules contains a group of fine-grained rules if we set $c \in \{0,1,\cdots,9\}$. Hence, the averaged results across the fine-grained ones are reported for each rule.

While these rules may not apply to every sample, they may provide an approximate estimation of a sample’s true posterior distribution. Table~\ref{tab:rule} presents the results for Kendall’s $\tau$ and the standard deviation of probabilities for non-ground-truth classes. The consistency of teacher outputs with the defined rules shows little variation; in the simpler Digits Five task, the predicted ranks are nearly identical (with Kendall’s $\tau$ approaching 1.0). However, larger teachers yield much smaller standard deviations than smaller ones, indicating less variability among non-ground-truth classes. These experiments on group classification further validate the initial observations in Sect.~\ref{sec:obser}.

\begin{table}[h]
    \caption{The consistency between the ranks predicted by different teachers with the defined rules and the standard deviation of the non-ground-truth class probability values.}
    \label{tab:rule}
    \begin{center}
        \begin{tabular}{>{\centering\arraybackslash}p{0.7cm}|>{\centering\arraybackslash}p{0.65cm}|>{\centering\arraybackslash}p{0.75cm}|>{\centering\arraybackslash}p{1.02cm}|>{\centering\arraybackslash}p{0.65cm}|>{\centering\arraybackslash}p{0.75cm}|>{\centering\arraybackslash}p{1.02cm}}
            \hline \hline
            \multirow{2}{*}{\tiny{CIFAR-100}} & \multicolumn{3}{c|}{Kendall's $\tau$} & \multicolumn{3}{c}{Standard Deviation} \\ \cline{2-7}
            & \tiny{ResNet20} & \tiny{WRN28-3} & \tiny{RNX29-64-4d} & \tiny{ResNet20} & \tiny{WRN28-3} & \tiny{RNX29-64-4d} \\ \hline \hline
            C.1 & 0.811 & 0.845 & 0.842 & 0.023 & 0.010 & 0.007 \\ \hline
            C.2 & 0.596 & 0.513 & 0.430 & 0.017 & 0.005 & 0.004 \\ \hline
            C.3 & 0.436 & 0.444 & 0.525 & 0.036 & 0.015 & 0.010 \\ \hline
            C.4 & 0.776 & 0.824 & 0.849 & 0.026 & 0.012 & 0.010 \\ \hline
            C.5 & 0.593 & 0.658 & 0.587 & 0.024 & 0.009 & 0.007 \\ \hline
            \rowcolor{gray!20} Avg & 0.643 & 0.657 & 0.647 & 0.025 & 0.010 & 0.008 \\
            \hline \hline
            \multirow{2}{*}{Digits} & \multicolumn{3}{c|}{Kendall's $\tau$} & \multicolumn{3}{c}{Standard Deviation} \\ \cline{2-7}
            & \tiny{VGG8} & \tiny{VGG13} & \tiny{VGG19} & \tiny{VGG8} & \tiny{VGG13} & \tiny{VGG19} \\ \hline \hline
            D.1 & 0.994 & 0.996 & 0.993 & 0.008 & 0.004 & 0.003 \\ \hline
            D.2 & 0.991 & 0.999 & 0.998 & 0.010 & 0.007 & 0.005 \\ \hline
            D.3 & 0.904 & 0.870 & 0.910 & 0.009 & 0.009 & 0.007 \\ \hline
            \rowcolor{gray!20} Avg & 0.963 & 0.955 & 0.967 & 0.009 & 0.007 & 0.005 \\
            \hline \hline
        \end{tabular}
    \end{center}
\end{table}

\section{Addressing Capacity Mismatch}
\noindent As introduced in Sect.~\ref{sec:bg}, the ``capacity mismatch'' refers to the phenomenon that larger networks may not teach students as well as smaller teachers~\cite{TAKD, KDEfficacy, ResKD-NC, ResKD-TIP, ATS}. The above observations show that the relative class affinity between the larger teacher and smaller teacher's outputs are basically consistent, while the absolute probabilities between non-ground-truth classes are not so discriminative for larger teachers. Hence, our proposed methods aim to enhance the distinctness between non-ground-truth class probabilities from different perspectives. The relative class affinity we observed ensures the correctness and effectiveness of our methods.

\subsection{Fusing Global Class Relations (FGCR)}
For each class $c$, we calculate the averaged probabilities via $\mathbb{E}_{\mathbf{x},y=c}[\mathbf{p}(\mathbf{x};\tau_0)]$, and advocate that this captures the global relationships among classes. According to Fig.~\ref{fig:cifar100-group}, the global relationships given by different teachers are similar in relative values, which may be useful for stable distillation. Hence, we fuse this to each training sample of the $c$-th class as follows:
\begin{equation}
    \hat{\mathbf{p}}(\mathbf{x};\tau) = (1.0 - \alpha)\mathbf{p}(\mathbf{x};\tau) + \alpha * \mathbb{E}_{\mathbf{x},y=c}[\mathbf{p}(\mathbf{x};\tau_0)], \label{eq:fgcr}
\end{equation}
where $\alpha$ is the hyper-parameter. The relative class affinity suggests that small teacher can be regarded as a form of label smoothing (LS) for the large teacher, as both ensure consistency in the rank order of classes' probabilities. Currently, the large teacher exhibits an overly one-hot output, necessitating LS to increase the variance of non-ground-truth class probabilities. Although LS can hurt knowledge KD~\cite{WhenLSHelp}, ~\cite{Revisit-LS-and-KD} demonstrates that label smoothing is compatible with KD at lower temperatures. Therefore, we set $\tau_0 < \tau$. As shown in Fig.~\ref{fig:cifar100-group}, the averaged probability distribution provided by larger networks still shows less varied values among non-ground-truth classes, which may limit the effectiveness of this method. Hence, we only provide this method as a trial, and it is not the focus of our paper. In fact, the performances of FGCR could surpass the distillation performance of a smaller teacher, while it is still worse than other advanced techniques as shown in Tab.~\ref{tab:compare-cap}.

\subsection{Regularizing Teachers (RegT)}
During the phase of training teachers, we add the following regularization to enhance the variance of non-ground-truth class probabilities:
\begin{equation}
    \ell(\mathbf{x},y) = \ell_{\text{CE}}(\mathbf{x},y) + \beta \left( \mathbf{p}_y(\mathbf{x};1.0) - \sigma(\mathbf{q}(\mathbf{x};1.0)) \right), \label{eq:regt}
\end{equation}
where $\mathbf{q}(\mathbf{x};1.0)=[\mathbf{p}_c(\mathbf{x};1.0)]_{c\neq y}$ and $\beta$ is the coefficient of the regularization term. Equation (\ref{eq:regt}) explicitly encourages teacher network to decrease the influence of the ground-truth class and enhance the distinctness of non-ground-truth classes. Different from other post-processing methods, this way could enhance the variance of non-ground-truth class probabilities in the stage of training teachers, which could provide an excellent initialization for distillation.

\subsection{Instance-Specific Asymmetric Temperature Scaling (ISATS)}
The previous work~\cite{ATS} proposes the approach named asymmetric temperature scaling (ATS) to tackle the capacity mismatch in KD. ATS applies different temperatures to the logits of ground-truth and non-ground-truth classes, i.e.,
\begin{eqnarray}
    &&\mathbf{p}_c(\tau_1, \tau_2) = \exp\left( \mathbf{f}_c / \tau_c \right)/\sum_{j\in [C]} \exp\left( \mathbf{f}_j / \tau_j \right), \label{eq:ats}
\end{eqnarray}
where $\tau_j = \tau_1$ for $j=y$ and $\tau_j=\tau_2$ for $j\neq y$. The recommended setting is $\tau_1 - \tau_2 \in [1, 2]$. This could effectively make the larger networks teach well again, but the effort in searching for proper hyper-parameters of $\tau_1$ and $\tau_2$ is huge.

\begin{table*}[t]
    \caption{Performance comparisons of addressing capacity mismatch in KD. The last three rows are our methods.}
    \label{tab:compare-cap}
    \begin{center}
        \begin{small}
            \begin{tabular}{l|c|c|c|c|c|c|c|c|c|c|c|c|c}
                \hline \hline
                Dataset & \multicolumn{3}{c|}{CIFAR-100}  & \multicolumn{3}{c|}{TinyImageNet} & \multicolumn{3}{c|}{CUB} & \multicolumn{3}{c|}{Stanford Dogs} & \multirow{3}{*}{Avg}  \\
                \cline{1-13}
                Teacher & \multicolumn{3}{c|}{ResNet110 (74.09)}  & \multicolumn{3}{c|}{WRN50-2(66.28)} & \multicolumn{3}{c|}{RNX101-32-8d (79.50)} & \multicolumn{3}{c|}{RNX101-32-8d (73.98)} &  \\
                \cline{1-13}
                Student & VGG8 & SFV1 & MV2 & ANet & SFV2 & MV2 & ANet & SFV2 & MV2 & ANet & SFV2 & MV2 & \\ \hline \hline
                NoKD & 69.92 & 70.04 & 64.75 & 34.62 & 45.79 & 52.03 & 55.66 & 71.24 & 74.49 & 50.20 & 68.72 & 68.67 & 60.51\\ \hline
                ST-KD & 72.30 & 73.22 & 66.56 & 36.16 & 49.59 & 52.93 & 56.39 & 72.15 & 76.80 & 51.95 & 69.92 & 72.06 & 62.50\\ \hline
                KD & 71.35 & 71.86 & 65.49 & 35.83 & 48.48 & 52.33 & 55.10 & 71.89 & 76.45 & 50.22 & 68.48 & 71.25 & 61.56\\ \hline
                ESKD & 71.88 & 72.02 & 65.92 & 34.97 & 48.34 & 52.15 & 55.64 & 72.15 & 76.87 & 50.39 & 69.02 & 71.56 & 61.74\\ \hline
                TAKD & {\bf 72.71} & 72.86 & 66.98 & 36.20 & 48.71 & 52.44 & 54.82 & 71.53 & 76.25 & 50.36 & 68.94 & 70.61 & 61.87\\ \hline
                SCKD & 70.38 & 70.61 & 64.59 & 36.16 & 48.76 & 51.83 & 56.78 & 71.99 & 75.13 & 51.78 & 68.80 & 70.13 & 61.41\\ \hline
                DIST & 71.22 & 71.10 & 65.01 & 35.95 & 45.75 & 50.56 & 48.88 & 71.79 & 75.34 & 49.85 & 67.69 & 68.97 & 60.18 \\ \hline 
                KD+ATS & 72.31 & 73.44 & {\bf 67.18} & {\bf 37.42} & {\bf 50.03} & 54.11 & 58.32 & 73.15 & 77.83 & 52.96 & 70.92 & {\bf 73.16} & 63.40 \\ \hline \hline
                KD+FGCR & 71.99 & 72.42 & 66.39 & 35.70 & 48.35 & 53.25 & 60.22 & 73.63 & {\bf 79.44} & 51.84 & 70.45 & 72.98 & 63.05 \\ \hline
                KD+RegT & 72.09 & 72.39 & 66.18 & 36.66 & 48.48 & 53.31 & 60.17 & 74.20 & 79.21 & 52.18 & 70.55 & 72.43 & 63.15 \\ \hline
                KD+ISATS & 72.46 & {\bf 73.80} & 67.04 & 37.15 & 49.61 & {\bf 55.49} & {\bf 61.73} & {\bf 74.23} & 78.84 & {\bf 54.36} & {\bf 70.98} & 72.73 & {\bf 64.04} \\ \hline \hline
            \end{tabular}
        \end{small}
    \end{center}
\end{table*}

As a step further solution, we extend ATS to Instance-Specific ATS (ISATS) that searches for proper temperatures for each training sample. Given a training sample $\mathbf{x}$, the predicted logit vector is $\mathbf{f}(\mathbf{x})$, and the optimal temperature that could enlarge the variance of softened non-ground-truth classes' probabilities is:
\begin{equation}
    \tau^\star(\mathbf{x}) = \arg\max_{\tau} v(\mathbf{q}(\mathbf{x};\tau)), \label{eq:ists}
\end{equation}
where $\mathbf{q}(\mathbf{x};\tau)$ denotes the probability vector of non-ground-truth classes in teacher network, i.e., $\mathbf{q}(\mathbf{x};\tau)=[\mathbf{p}_c(\mathbf{x};\tau)]_{c\neq y}$ and $\mathbf{p}(\mathbf{x};\tau)=\text{SF}(\mathbf{f}(\mathbf{x});\tau)$. It is noted that, according to relative class affinity, enlarging the teacher model does not alter the rank order among non-ground-truth classes. Consequently, all non-target classes share the same $\tau$, eliminating the need to apply different $\tau$ values across classes. $v(\cdot)$ calculates the variance of the elements in a vector. That is, the optimal temperature for each instance is searched to make the probabilities of non-ground-truth classes more distinct. Then, we set $\tau_1(\mathbf{x}) = \tau^\star + 1$ and $\tau_2(\mathbf{x}) = \tau^\star$ as recommended by ATS~\cite{ATS}. Equation (10) has no closed-form solution. So like ATS, ISATS still requires manual search to ensure an appropriate temperature. Nevertheless, ISATS has two advantages over ATS. First, ISATS saves the effort of finding the proper value of $\tau_2$ and $\tau_1$. ATS requires performing multiple distillation processes and comparing the experimental results at different temperatures to determine proper $\tau$. In contrast, ISATS can directly compute $\tau^\star$ based on (\ref{eq:ists}) in advance, eliminating the need for comparative distillation experiments. Second, ISATS considers a more fine-grained way to release the discriminative information contained in wrong classes for each training sample. In summary, ISATS is an all-around improvement over ATS.

\section{Performance Comparisons}
\noindent This section provides performance comparisons of addressing capacity mismatch in KD.
The experimental studies follow the settings in~\cite{ATS} and utilize the publicly available code~\footnote{\url{https://github.com/lxcnju/ATS-LargeKD}}. We directly cite some experimental results for comparisons. Specifically, the utilized datasets are CIFAR-10/CIFAR-100~\cite{cifar}, TinyImageNet~\cite{TinyImageNet}, CUB~\cite{CUB}, and Stanford Dogs~\cite{dogs}. We choose teacher models significantly larger than their corresponding student models as we do in Section 4. Teacher networks include versions of ResNet~\cite{ResNet}, WideResNet~\cite{WideResNet}, and ResNeXt~\cite{ResNeXt}. Student networks are VGG~\cite{VGG}, ShuffleNetV1/V2~\cite{ShuffleNet, ShuffleNetV2}, AlexNet~\cite{AlexNet}, and MobileNetV2~\cite{MobileNetV2}.
We train networks on corresponding datasets for 240 epochs. SGD optimizer with a momentum value of 0.9 is used. The learning rate is 0.05 by default, and the batch size is 128. 
For our proposed FGCR, we set $\tau_0 = \tau - 1$ and $\alpha \in \{0.1, 0.5\}$, and the best results are reported. For RegT, $\beta$ is searched in $\{0.01, 0.001\}$. For ISATS, we search for the best temperature for each instance in the scope of $\{1.0, 2.0, 3.0, 4.0, 5.0, 6.0, 8.0\}$. The distribution of the optimal instance-specific temperatures on CIFAR-100 is as follows: 4.0 accounts for about $31\%$, 5.0 accounts for about $27\%$, 3.0 accounts for about $22\%$, and others account for $20\%$.

\begin{table*}[t]
    \caption{Ensemble performance comparisons of addressing capacity mismatch in KD. The last three rows are our methods.}
    \label{tab:compare-cap-ens}
    \begin{center}
        \begin{small}
            \begin{tabular}{l|c|c|c|c|c|c|c|c|c|c|c|c|c}
                \hline \hline
                Dataset & \multicolumn{3}{c|}{CIFAR-100}  & \multicolumn{3}{c|}{TinyImageNet} & \multicolumn{3}{c|}{CUB} & \multicolumn{3}{c|}{Stanford Dogs} & \multirow{3}{*}{Avg}  \\
                \cline{1-13}
                Teacher & \multicolumn{3}{c|}{ResNet110 (74.09)}  & \multicolumn{3}{c|}{WRN50-2(66.28)} & \multicolumn{3}{c|}{RNX101-32-8d (79.50)} & \multicolumn{3}{c|}{RNX101-32-8d (73.98)} &  \\
                \cline{1-13}
                Student & VGG8 & SFV1 & MV2 & ANet & SFV2 & MV2 & ANet & SFV2 & MV2 & ANet & SFV2 & MV2 & \\ \hline \hline
                NoKD Ens & 72.77 & 73.61 & 67.76 & 39.37 & 50.69 & 56.40 & 59.84 & 74.43 & 77.47 & 54.04 & 71.65 & 72.53 & 64.21\\ \hline
                ResKD & 73.89 & {\bf 76.03} & 69.00 & 38.66 & 51.93 & 57.32 & 62.60 & 75.29 & 76.27 & 54.68 & 70.73 & 72.85 & 64.94\\ \hline
                KD+ATS+Ens & 74.86 & 75.05 & {\bf 69.50} & {\bf 40.42} & 52.14 & 58.47 & 62.00 & 76.26 & 78.97 & 55.69 & 73.22 & 74.67 & 65.94 \\ \hline \hline
                KD+FGCR+Ens & 73.69 & 74.76 & 68.22 & 39.20 & 52.71 & 56.44 & 61.94 & {\bf 76.84} & 80.37 & 54.25 & 72.70 & 74.83 & 65.50 \\ \hline
                KD+RegT+Ens & {\bf 74.88} & 74.95 & 68.45 & 39.37 & {\bf 53.78} & 57.35 & 62.11 & 75.64 & 80.82 & 55.03 & 72.73 & 74.25 & 65.78  \\ \hline
                KD+ISATS+Ens & 74.55 & 74.43 & 68.61 & 40.25 & 53.18 & {\bf 58.78} & {\bf 63.82} & 75.07 & {\bf 80.87} & {\bf 56.62} & {\bf 73.53} & {\bf 75.30} & {\bf 66.25} \\
                \hline \hline
            \end{tabular}
        \end{small}
    \end{center}
\end{table*}

\subsection{Performance Comparisons}
We compare with SOTA methods and list the results on CIFAR-100, TinyImageNet, CUB, and Dogs in Tab.~\ref{tab:compare-cap}. The compared methods include ESKD~\cite{KDEfficacy}, TAKD~\cite{TAKD}, SCKD~\cite{SCKD}, and ATS~\cite{ATS}. NoKD trains students without the teacher's supervision. ST-KD trains students under the guidance of a smaller teacher. KD trains students under the guidance of the larger teacher. The larger teachers are ResNet110, WRN50-2, RNX101-32-8d, and RNX101-32-8d for the four datasets, while the smaller teachers are ResNet20, WRN26-2, RNX50-32-4d, and RNX50-32-4d, respectively. The values in ``()'' display the test accuracy of the large teachers. The last column of the table shows the average performance of corresponding rows. The last three rows present the performances of our proposed algorithms. These methods could improve the performances of students when taught by a larger network and surpass that of KD and ST-KD, which verifies that enhancing the variance of non-ground-truth classes could indeed make larger networks teach well again. Although dataset bias exists in a few circumstances, the proposed ISATS could achieve the SOTA results when compared with ATS. It is easy to understand that ISATS searches for optimal temperatures in an instance-specific manner, which could enhance the variance of non-ground-truth class probabilities more effectively. The two methods presented in Sections 5.1 and 5.2 are also meaningful. RegT belongs to the pre-processing method that enhances the variance of non-ground-truth classes during the phase of training teachers, which also presents good performances. RegT does not directly affect the teacher model's error class variance like ATS, so its performance is not as good as ATS. However, together with ESKD \cite{KDEfficacy}, it indicates that a more suitable teacher is benefical to KD performance improvement and provides a simple and effective method for teacher regularization. Similarly, the performance improvement of FGCR is less pronounced than ATS, but it indirectly supports the rationale for moderate LS. The experimental data for these two methods, from the perspective of wrong-class variance that we introduced, demonstrate the validity of previous works that tried to address capacity mismatch.

We also verify the ensemble performances after distillation because ResKD~\cite{ResKD-NC, ResKD-TIP} improves the students' performances by introducing the residual student and taking the two residual students' ensemble. Although ResKD surpasses the performance of ATS, this comparison is not fair. Hence, we also provide ensemble performances of two separately trained students under different initialization and training seeds. For a fair comparison, we also test the performances of our methods by repeating the corresponding algorithms two times and making predictions via the ensemble of the obtained two student networks. The results are listed in Tab.~\ref{tab:compare-cap-ens}. Clearly, our proposed ISATS could also achieve better performances when compared with ATS and other methods.

\begin{figure}[tbp]
    \centering
    \includegraphics[width=\linewidth]{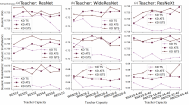}
    \caption{Distillation results via TS, ATS, and ISATS on CIFAR-100. (a)ResNet teachers;(b)WideResNet teachers;(c)ResNeXt teachers}
    \label{fig:compare-cap}
\end{figure}

\subsection{Comparing ISATS with TS and ATS}
\noindent We then especially focus on the comparison of ISATS with TS and ATS. Specifically, we follow the experimental settings in~\cite{ATS} and additionally plot the performance curves when utilizing our proposed ISATS. Fig.~\ref{fig:compare-cap} shows several pairs of KD experimental studies, which include the combinations of three series of teacher networks (i.e., ResNet, WideResNet, and ResNeXt) and three student networks (i.e., VGG8, ShuffleNetV1, and MobileNetV2). The utilized dataset is CIFAR-100. Each plot shows three curves, including TS (i.e., KD), ATS (i.e., KD+ATS), and ISATS (i.e., KD+ISATS). The x-axis represents the increasing capacity of teachers. The curves of TS imply that the student's performance taught by teacher networks becomes worse when the teacher capacity increases, i.e., the capacity mismatch phenomenon. ATS could mitigate this phenomenon and make larger teachers teach well or better again. Our proposed ISATS has the same effect and could surpass ATS in most cases.	

Fig.~\ref{fig:compare-cap} further demonstrates that ISATS can achieve superior KD performance with a given teacher model size, avoiding the substantial effort required to find an appropriate teacher model in the situation of TS. Notably, ISATS outperforms ATS in distillation performance, particularly when the teacher model is large.

\begin{figure}[tbp]
    \centering
    \includegraphics[width=\linewidth]{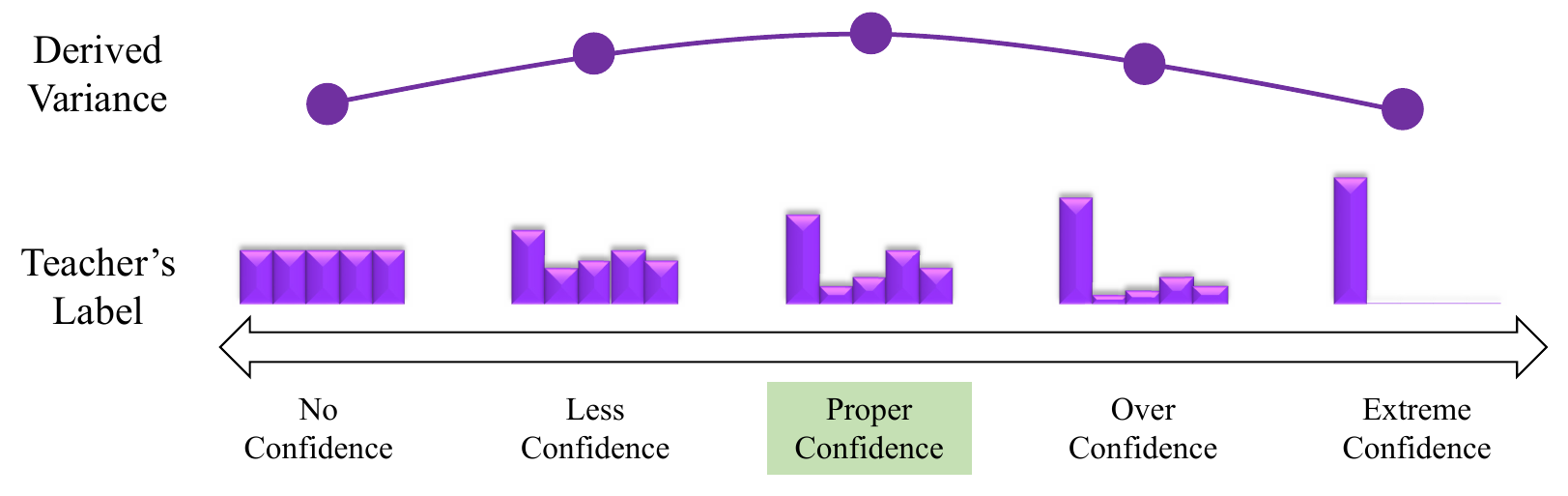}
    \caption{The teacher's label under different levels of confidence. The dark knowledge provided by the teacher with proper confidence is preferred for reaching the highest derived variance (variance of probabilities of non-ground-truth classes).}
    \label{fig:diss}
\end{figure}

\section{Discussion} \label{sec:dis}
\noindent This paper explores the influence of teachers' capacity on their provided distillation labels, i.e., the dark knowledge in KD. A larger teacher may be over-confident and provide probability vectors that are less discriminative between non-ground-truth classes. If we view the capacity as an influencing factor of the confidence level, then we could abstract the illustration shown in Fig.~\ref{fig:diss}. In fact, the confidence level is related to the following factors.
\begin{itemize}
    \item \textbf{Network Capacity}. This paper mainly explores this factor. An extremely smaller network tends to provide uniform predictions, while an extremely larger network provides ``one-hot" ones.
    \item \textbf{Training Process}. With the training process of a network, it becomes more and more confident. In the beginning, it is ignorant and provides uniform predictions. After enough training steps, it may output ``one-hot" vectors. That is, a proper training level of teachers also matters a lot in the KD process. Some works have pointed out that intermediate model checkpoints could be better teachers than the fully converged model~\cite{CheckpointKD}.
    \item \textbf{Temperature Factor}. Given a trained network, we could also adjust the temperature and obtain different types of probability vectors. If we take a very high temperature, the distribution becomes uniform. In contrast, a lower temperature leads to ``one-hot" results. Consequently, traditional KD methods do not use a too-small or too-large temperature~\cite{KD}.
\end{itemize}

As shown in Fig.~\ref{fig:diss}, a proper confidence level could generate a rational probability distribution that facilitates the KD process. Our work points out why the teacher with high capacity does not perform well in KD and proposes several solutions to adjust their over-confident outputs.

\section{Conclusion}
\noindent This paper studies the dark knowledge in KD under various teacher capacities. Two observations are first presented, i.e., the probability vectors provided by larger teachers are less distinct among non-ground-truth classes, while the relative probability values seem consistent among different teachers. Abundant experimental studies are provided to explain these two observations. The less varied non-ground-truth class probabilities make the student hardly grasp the absolute affinities of non-ground-truth classes to the target class, leading to the interesting ``capacity mismatch'' phenomenon in KD. Multiple effective and simple solutions are proposed to solve this problem. Finally, a unified perspective about dark knowledge under various confidence levels is provided for future studies.

\Acknowledgements{This work is supported by National Science and Technology Major Project (Grant No. 2022ZD0114805).}

\bibliography{fcs}
\bibliographystyle{fcs}

%\bibliography{paper}
%\bibliographystyle{fcs}

%% \Biography{#1}{#2}
%% #1 is the file name of photo,
%% #2 is author's biography

\Biography{fanws}{Wen-Shu Fan received his B.Sc. degree in School of Computer Science, Northwestern Polytechnical University in June 2019. He is currently working towards a Ph.D. degree with the School of Artificial Intelligence in Nanjing University, China. His research interests include knowledge distillation, model reuse and deep learning.}

\Biography{lixc}{Xin-Chun Li received his M.Sc. degree with the National Key Lab for Novel Software Technology, the Department of Computer Science and Technology in Nanjing University, China. He is working towards a Ph.D. degree with the School of Artificial Intelligence in Nanjing University, China. Up until now, he has published over 10 papers in national and international journals or conferences such as NeurIPS, CVPR, KDD, ICASSP, SCIS, etc. His research interests lie primarily in machine learning and data mining, including federated learning.}

\Biography{zhandc}{De-Chuan Zhan joined in the LAMDA Group on 2004 and received his Ph.D. degree in Computer Science from Nanjing University in 2010, and then serviced in the Department of Computer Science and Technology of Nanjing University as an Assistant Professor from 2010, and as an Associate Professor from 2013. Then he joined the School of Artificial Intelligence of Nanjing University as a Professor from 2019. His research interests mainly include machine learning and data mining, especially working on mobile intelligence, distance metric learning, multi-modal learning, etc. Up until now, he has published over 60 papers in national and international journals or conferences such as TPAMI, TKDD, TIFS, TSMSB, IJCAI, ICML, NIPS, AAAI, etc.}

\end{document}